\documentclass[10pt,twocolumn,letterpaper]{article}
\usepackage[letterpaper,textwidth=6.875in,textheight=8.875in,top=0.9in,columnsep=0.3125in]{geometry}
\usepackage{times}
\usepackage[table,dvipsnames]{xcolor}
\usepackage{amsmath,amssymb,booktabs,graphicx,array,tabularx}
\usepackage{algorithm,algpseudocode}
\usepackage[numbers,sort&compress]{natbib}
\usepackage[font=small,labelsep=period]{caption}
\usepackage{enumitem}
\setlist[itemize]{leftmargin=*,nosep}
\usepackage{titlesec}
\titleformat{\section}{\large\bfseries}{\thesection.}{.4em}{}
\titleformat{\subsection}{\normalsize\bfseries}{\thesubsection.}{.4em}{}
\titlespacing*{\section}{0pt}{1.5ex plus .5ex}{.8ex}
\titlespacing*{\subsection}{0pt}{1.3ex plus .5ex}{.6ex}
\renewenvironment{abstract}{\par\begin{center}\large\bfseries Abstract\end{center}\itshape\noindent\ignorespaces}{\par\medskip}
\usepackage{microtype}
\newcommand{\om}{VGGT-$\Omega$}
\newcommand{\pii}{$\pi^3$}

\newcommand{\paperVenue}[2]{\nobreak\hspace{.24em}{\scriptsize\textcolor{black!62}{#1~'#2}}}

\usepackage[hidelinks,pdfauthor={Shuang Yang, Liangyu Xiang, Yufan Hu, Quan Fang, Junyu Gao},pdftitle={MoDiR: Motion-Aware Compositional Dictionaries for Efficient Feed-Forward 4D Reconstruction}]{hyperref}
\title{\Large\bfseries MoDiR: Motion-Aware Compositional Dictionaries\\for Efficient Feed-Forward 4D Reconstruction}
\author{\large\normalfont Shuang Yang$^{1,2}$\quad Liangyu Xiang$^{1}$\quad Yufan Hu$^{3}$\quad Quan Fang$^{4}$\quad Junyu Gao$^{1\dagger}$\\[4pt]
\large $^1$Institute of Automation, Chinese Academy of Sciences\\
\large $^2$University of Chinese Academy of Sciences\\
\large $^3$University of Science and Technology Beijing\\
\large $^4$Beijing University of Posts and Telecommunications\\[3pt]
\small\texttt{yangshuang21a@mails.ucas.ac.cn, xiangliangyu2023@ia.ac.cn}\\
\small\texttt{yufanhu@ustb.edu.cn, qfang@bupt.edu.cn, junyu.gao@nlpr.ia.ac.cn}}

\date{}
\begin{document}
\raggedbottom
\maketitle
\begingroup
\renewcommand{\thefootnote}{\textdagger}
\footnotetext{Corresponding author.}
\endgroup
\begin{abstract}
Recent feed-forward 4D reconstruction models recover 3D scene geometry and motion directly from video, but their reliance on global attention to combine information across frames makes long sequences costly to process. In multi-view 3D reconstruction, token merging reduces this cost by assigning tokens with similar features to a shared representative. However, in dynamic videos, we observe that patches with similar features can correspond to different 3D positions and that tokens matched to the same representative can produce substantially different attention outputs. Based on these observations, we propose MoDiR, a motion-aware token reduction method that selects representative tokens and combines their attention outputs using token-specific weights. MoDiR samples spatial representatives and uses cross-frame motion cues to select temporal representatives, then combines the two sets to form the dictionary. MoDiR then fits the remaining token features as sparse convex combinations of dictionary entries and reuses the fitted weights to approximate global attention outputs for selected tokens that pass the fit test. Extensive experiments across multiple feed-forward reconstruction models and datasets show a favorable trade-off between accuracy and efficiency in geometry estimation and 3D point tracking, with up to a $4.3\times$ inference speedup for long-sequence geometry estimation.
\end{abstract}

\section{Introduction}
\label{sec:intro}

Early feed-forward visual geometry models jointly estimated camera parameters and dense 3D geometry from multiple views in a single forward pass~\cite{r19,yang2025fast3r,zhang2025flare}. Extending this approach to dynamic scenes also requires estimating how the 3D positions of observed points change over time~\cite{vdpm,huang2026fastspatial,xiao2025spatialtrackerv2}. Recent feed-forward 4D reconstruction models~\cite{r25,fourrc,vdpm} process multiple frames jointly, using the resulting temporal context to recover 3D scene geometry and motion directly from video. As input sequences grow longer, however, the computational cost of joint processing rises rapidly~\cite{r10,r18}.

\begin{figure*}[t]
\centering\includegraphics[width=\textwidth]{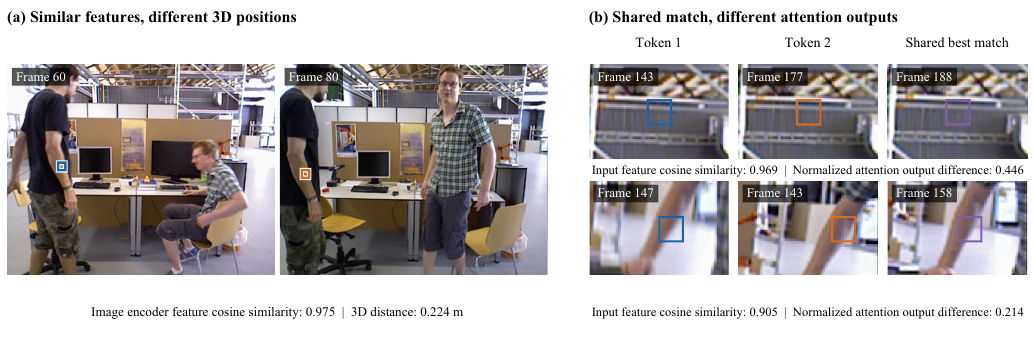}
\caption{Feature similarity, 3D positions, and attention outputs in dynamic video. (a) The marked patches in Frames 60 and 80 have a cosine similarity of 0.975 in the features extracted by \om{}'s DINOv3-initialized image encoder~\cite{r20}, yet their reference 3D positions are 0.224 m apart in a common coordinate system. (b) Each row shows two patch tokens (blue and orange) and their shared best match (purple). In global attention layer 12 of \om{}, the same candidate has the highest input feature cosine similarity to each token within its respective candidate set. Copying that candidate's attention output to both tokens would give them identical outputs, although their original outputs differ substantially. The normalized attention output difference is the Euclidean distance between the two projected outputs before residual addition, divided by the root mean square of their norms. Appendix~\ref{app:diagnostics} gives the example selection and measurement details.}
\label{fig:motivation}
\end{figure*}

Feed-forward visual geometry models use global attention~\cite{r19,r25,fourrc} to establish geometric relationships across frames. With a fixed number of tokens per frame, the computational cost of dense global attention grows quadratically with the number of frames~\cite{r10,r18,vaswani2017attention}. A natural way to reduce this cost is to remove redundant tokens with highly similar features from the sequence passed to global attention. In multi-view 3D reconstruction, token merging methods do this by merging selected tokens into their most similar representatives~\cite{r12}, then copying each representative's attention output back to the positions of the tokens merged into it~\cite{r21}. However, applying this rule directly to 4D reconstruction can group tokens from patches that have similar features but correspond to different 3D positions at their respective observation times and assign a common attention output to tokens whose original outputs differ substantially. In Fig.~\ref{fig:motivation}(a), two image patches have a cosine similarity of 0.975 in the features extracted by \om{}'s DINOv3-initialized image encoder~\cite{r20,simeoni2026dinov3}, yet their 3D positions are 0.224 m apart in a common reference frame. In Fig.~\ref{fig:motivation}(b), this token merging strategy would merge both tokens into the same representative because it is the most similar candidate for each, even though their outputs in the original global attention layer differ substantially.

These observations motivate a token reduction strategy for feed-forward 4D reconstruction that accounts for changes in scene geometry when selecting representatives and allows tokens that share representatives to receive different approximations of their global attention outputs. To this end, we propose MoDiR, a motion-aware token reduction method that combines spatial and temporal representatives to form a dictionary and uses token-specific weights to combine the representatives' attention outputs. Specifically, MoDiR samples spatial representatives to cover each frame and uses cross-frame motion cues to identify candidates for temporal representatives. It selects temporal representatives from candidates whose features are poorly approximated by sparse convex combinations of the available representatives. Once the dictionary is formed, MoDiR uses the same fitting procedure to represent the remaining token features and reuses the fitted weights to approximate global attention outputs for selected tokens that pass the fit test.

Our contributions are summarized as follows.

\begin{itemize}
\item We show that in dynamic videos, merging by feature similarity can group tokens from patches at different 3D positions. Copying a representative's attention output gives all tokens in its group the same output, even when their original outputs differ substantially. We therefore use cross-frame motion cues to select temporal representatives and token-specific weights to approximate attention outputs.
\item We propose MoDiR, a motion-aware token reduction method that uses sparse convex fitting to select temporal representatives that complement the spatial representatives. MoDiR combines both sets into a dictionary, fits the remaining token features as sparse convex combinations of dictionary entries, and reuses the fitted weights to approximate global attention outputs for selected tokens that pass the fit test.
\item We demonstrate a favorable trade-off between accuracy and efficiency in geometry estimation and 3D point tracking across multiple feed-forward reconstruction models and datasets, with up to a $4.3\times$ inference speedup for geometry estimation on long sequences.
\end{itemize}

\section{Related Work}

\subsection{Feed-Forward Visual Geometry Transformers}

Feed-forward visual geometry Transformers have progressed from multi-view 3D reconstruction to the recovery of scene geometry and motion from video~\cite{chen2026dggt}. VGGT~\cite{r19} unifies camera and dense geometry prediction in a single forward pass. Subsequent models refine multi-view inference through a permutation-equivariant architecture~\cite{r22} and a unified depth--ray prediction target~\cite{da3}. Beyond static multi-view reconstruction~\cite{wang2024dust3r}, \om{}~\cite{r20} and related models~\cite{r26,r27,r6,cong2026flow3r} estimate geometry in dynamic scenes, and explicit motion prediction has developed through dynamic point maps~\cite{dpm,vdpm,dynvggt}, scene flow~\cite{r9,point4cast}, and trajectory fields~\cite{traceanything,omnix}. D4RT~\cite{r25} and 4RC~\cite{fourrc} encode the video once and reuse the resulting representation to predict 3D positions or displacements at requested times. However, visual geometry Transformers use global attention~\cite{r19,r25,fourrc} to relate tokens across frames, and the quadratic cost of these interactions~\cite{r18,r5} limits the length of sequences they can process efficiently. A natural objective is therefore to reduce the number of tokens involved in global attention while preserving geometric fidelity.

\begin{figure*}[t]
\centering\includegraphics[width=\textwidth]{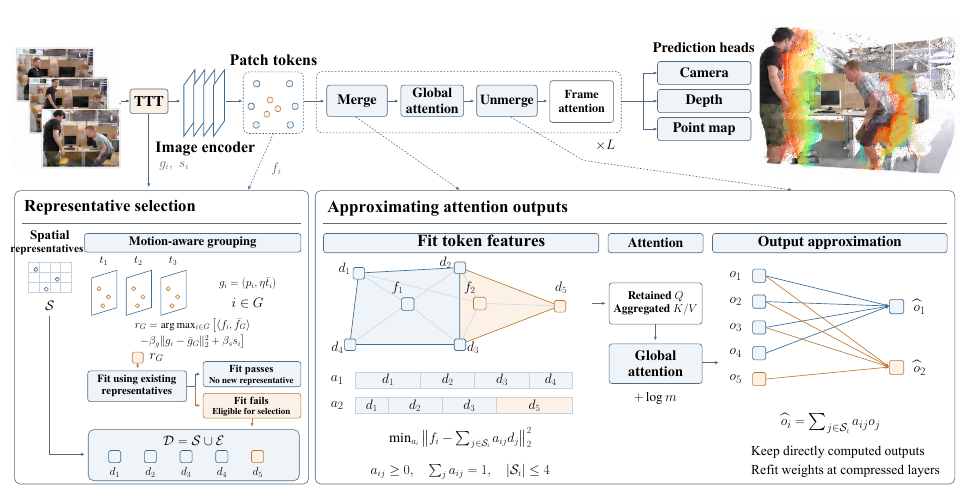}
\caption{\textbf{MoDiR at a global attention layer.} On the left, cross-frame motion cues from the TTT block identify candidates for temporal representatives. Candidates whose features are poorly approximated by the available representatives are eligible for selection. The spatial and selected temporal representatives form the dictionary. On the right, MoDiR fits the remaining patch tokens to the dictionary. For each token whose query is omitted, it reuses the fitted weights to combine representative attention outputs and places the approximation at the token's original position.}
\label{fig:method}
\end{figure*}

\subsection{Efficient Feed-Forward 4D Reconstruction}

Prior work has improved the efficiency of feed-forward visual geometry Transformers through token reduction~\cite{r10,r12,r3,r21}, sparse~\cite{r18,r5} or compressed global attention~\cite{r15,r23}, and long-sequence inference~\cite{stac,cabon2025must3r} with compact scene representations~\cite{vggt3,zipmap,wang2025cut3r} or input partitioning~\cite{merg3r}. Among these approaches, token merging offers a natural and direct way to accelerate multi-view reconstruction by combining redundant tokens from overlapping views into a shorter sequence for global attention. FastVGGT~\cite{r10} samples representatives across image regions to maintain spatial coverage and copies each merged token's attention output back to the positions it represents. LiteVGGT~\cite{r12} uses geometric importance to guide token selection, and Co-Me~\cite{r3} uses predicted confidence to decide which groups to merge. HTTM~\cite{r21} matches tokens within spatiotemporal blocks and uses separate assignments for each attention head, preserving token diversity after head concatenation. However, sharing representatives raises two challenges in feed-forward 4D reconstruction. Matching by feature similarity can group tokens from patches at different 3D positions. When a representative's output is copied, its group shares that output even when the tokens' original global attention outputs differ substantially. To address these issues, MoDiR selects temporal representatives using cross-frame motion cues and approximates global attention outputs using token-specific fitting weights.

\section{Method}\label{sec:method}

\subsection{Token Reduction in Dynamic Scenes}

Feed-forward 4D reconstruction models recover 3D scene geometry and motion directly from video. The image encoder converts local regions of each frame into patch tokens. The Transformer backbone updates the token features by alternating between global attention, which combines information across frames, and frame attention, which exchanges information within each frame. Prediction heads or decoders use the resulting features to estimate camera parameters, depth maps, 3D point maps, or 3D point trajectories, depending on the task. Because global attention processes tokens from all frames jointly, its computational cost grows rapidly with sequence length.

Token merging shortens the sequence passed to global attention, and the resulting attention outputs are then mapped back to the original token positions. The choice of representatives determines which observations the layer can attend to separately. When tokens from different frames are merged, the layer attends to their combined feature. This can suppress differences that help the model recover changes in scene geometry. Copying the representative's attention output gives every token in the group the same output, even when the original layer would produce substantially different outputs for them. Errors can accumulate as subsequent layers operate on the features updated with these approximate outputs, affecting the final reconstruction.

These two limitations motivate motion-aware representative selection  and token-specific output approximation .

\subsection{Motion-Aware Representative Selection}\label{sec:representatives}

\paragraph{Spatial representatives.}
We retain all patch tokens from the first frame and sample one from each nonoverlapping $2\times2$ region in subsequent frames~\cite{r10}. These tokens form the set of spatial representatives $\mathcal S$. To find observations they may represent poorly, we use cross-frame motion cues to propose candidates for temporal representatives.

\paragraph{Motion cues and candidate grouping.}
The frozen backbone estimates 3D geometry jointly for a small subset of the input frames, which we use as anchor frames. We propagate the estimates to intervening frames using bidirectional optical flow. We derive a confidence score $c_i$ for each patch from forward--backward consistency. The score is higher when following the flow in both directions returns the patch closer to its starting position. Using reliable correspondences and the anchor predictions, we train a lightweight adapter~\cite{guo2026phaselora} at test time~\cite{r16} to refine the propagated 3D positions. Figure~\ref{fig:method} labels this preparation TTT for test-time training.

We use a score $s_i$ to identify patches for grouping. It combines changes in 3D position between corresponding patches, local variation in these changes and in optical flow, and the term $1-c_i$, which is larger for less reliable correspondences. We group the selected patches by 3D position and short time intervals. For patch $i$, the geometry--time descriptor $g_i=(p_i,\eta\bar t_i)$ combines its refined 3D position $p_i$ and normalized observation time $\bar t_i$, with $\eta=0.25$. The indices $t_1,t_2,t_3$ in Fig.~\ref{fig:method} denote the frames in the grouping example.

Each group $G$ proposes one token. We base this choice in part on the current features of its tokens. Let $z_i$ denote the feature of token $i$ after the backbone's normalization at the input to global attention. We use

\begin{equation}\label{eq:representation}
f_i=\frac{z_i}{\max(\lVert z_i\rVert_2,\epsilon)},
\end{equation}

where $\epsilon$ is a small positive constant that prevents division by zero. We omit the layer index from the features for clarity. The candidate $r_G$ is the token index that maximizes the following score:

\begin{equation}\label{eq:representative-score}
\begin{aligned}
r_G=\arg\max_{i\in G}\bigl\{&\langle f_i,\bar f_G\rangle-\beta_g\lVert g_i-\bar g_G\rVert_2^2\\
&+\beta_s s_i\bigr\}.
\end{aligned}
\end{equation}

Here $\bar f_G$ is the normalized mean feature and $\bar g_G$ is the mean geometry--time descriptor of the group. The first term favors features similar to the group's mean. The second penalizes distance from the mean descriptor, and the third favors a higher $s_i$. We use $\beta_g=0.01$ and $\beta_s=0.005$.

\paragraph{Selecting temporal representatives.}
We test whether the available representatives can approximate each candidate's feature. A fit uses at most $K=4$ representatives. Let $\mathcal S_i$ index the representatives used to fit token $i$, and let $d_j=f_j$ be the normalized feature of representative $j$. Given $\mathcal S_i$, we fit the weights $a_i=(a_{ij})_{j\in\mathcal S_i}$ by minimizing the squared feature error:

\begin{equation}\label{eq:fit}
\begin{gathered}
\min_{a_i}\left\lVert f_i-\sum_{j\in\mathcal S_i}a_{ij}d_j\right\rVert_2^2,\\
a_{ij}\geq0,\qquad
\sum_{j\in\mathcal S_i}a_{ij}=1,\qquad
|\mathcal S_i|\leq K.
\end{gathered}
\end{equation}

The fitted feature $\hat f_i=\sum_{j\in\mathcal S_i}a_{ij}d_j$ is a \emph{sparse convex combination}. It combines at most $K$ representative features with nonnegative weights that sum to one. The fit passes when

\begin{equation}\label{eq:accept}
\begin{aligned}
\cos(f_i,\hat f_i)&\geq\tau_l,\\
\lVert f_i-\hat f_i\rVert_2&\leq\sqrt{2(1-\tau_l)},\\
\lVert\hat f_i\rVert_2&\geq\rho_{\min}.
\end{aligned}
\end{equation}

The threshold $\tau_l$ at layer $l$ controls the required cosine similarity and the maximum feature error. The minimum norm $\rho_{\min}$ rejects near-zero fitted features.

We first fit candidates using the spatial representatives. At each later selection, we fit all candidates using the representatives available at the start of that selection. Candidates whose fits fail are eligible to become temporal representatives. We rank them by the product of $s_i$ and the feature error $\lVert f_i-\hat f_i\rVert_2$ and select within the remaining budget. The spatial representatives $\mathcal S$ and selected temporal representatives $\mathcal E$ form the dictionary $\mathcal D=\mathcal S\cup\mathcal E$. Appendix~\ref{app:implementation} details geometry preparation, grouping, retrieval, and fitting.

\subsection{Token-Specific Output Approximation}\label{sec:approximation}

Once the dictionary is formed, we use the same fitting procedure for each remaining patch token. We choose its representatives, fit their weights using Eq.~\eqref{eq:fit}, and check the fit using Eq.~\eqref{eq:accept}. Dictionary representatives and patch tokens whose fits fail retain their queries and receive directly computed attention outputs. In selected layers of \pii{}, we also retain queries with high fitting residuals (Appendix~\ref{app:implementation}). For the other patch tokens whose fits pass, we omit their queries and use their fitted weights to combine representative attention outputs.

\paragraph{Merging keys and values.}
In self-attention, a query is compared with the keys to determine how much each corresponding value contributes to the output. The label \emph{Retained Q} in Fig.~\ref{fig:method} refers to the queries for which an output is computed directly. We compute keys and values at all input positions, including those whose queries are omitted, and apply the backbone's normalization and positional encoding where applicable. The merged sequence contains one key--value pair for each dictionary representative, each patch token whose fit failed, and each special token.

Let $R_{ij}$ be the weight with which token $i$ contributes its key and value to pair $j$. Tokens that have their own pair contribute to it with unit weight. To limit the cost of merging, each other token contributes to at most two representatives using its largest fitted weights, renormalized to sum to one. For one attention head, let $k_i$ and $v_i$ be the key and value of token $i$ before merging. The total weight $m_j$ assigned to pair $j$ and its merged key and value are

\begin{equation}\label{eq:merge}
\begin{gathered}
m_j=\sum_iR_{ij},\\
\bar k_j=\frac{1}{m_j}\sum_iR_{ij}k_i,\qquad
\bar v_j=\frac{1}{m_j}\sum_iR_{ij}v_i.
\end{gathered}
\end{equation}

The $+\log m$ term in Fig.~\ref{fig:method} is the proportional attention correction~\cite{r2}. It adds $\log m_j$ to the attention score for pair $j$ to account for its total weight. For a patch token with a retained query $q_i$, the output of one head is

\begin{equation}\label{eq:attention}
h_i=\sum_j\operatorname{softmax}_j\!\left(
\frac{q_i^\top\bar k_j}{\sqrt{d_h}}+\log m_j
\right)\bar v_j,
\end{equation}

where $d_h$ is the key dimension in one head. Special tokens retain their queries and attend to the original keys and values. Additional queries retained for patch tokens use the same merged keys and values.

\paragraph{Approximating attention outputs.}
Let $Q$ be the set of indices of tokens whose queries are retained. For $j\in Q$, let $o_j$ be the attention output after concatenating the heads and applying the backbone's output projection. We keep each directly computed output at its original position. For a token whose query is omitted, we combine the attention outputs of its representatives using all its fitted weights:

\begin{equation}\label{eq:unmerge}
\hat o_i=
\begin{cases}
o_i, & i\in Q,\\[3pt]
\displaystyle\sum_{j\in\mathcal S_i}a_{ij}o_j, & i\notin Q.
\end{cases}
\end{equation}

In Fig.~\ref{fig:method}, the fits for $f_1$ and $f_2$ share the representatives with features $d_1,d_2,d_3$. The first fit also uses $d_4$, and the second uses $d_5$. Each token reuses its fitted weights, $a_1$ or $a_2$, to combine its representatives' attention outputs and obtain $\hat o_1$ or $\hat o_2$. We place all outputs at the original token positions before frame attention. The prediction heads use the final backbone features to produce the camera and dense geometry outputs shown in Fig.~\ref{fig:method}.

\subsection{Cross-Layer Reuse and Computational Cost}\label{sec:cost}

The $\times L$ in Fig.~\ref{fig:method} denotes $L$ repetitions of the backbone attention block. Across the global attention layers where MoDiR is applied, we reuse the spatial representatives and select temporal representatives periodically. At each such layer, we refit the weights using the current features. If a token's previous representatives are still available and the new fit passes Eq.~\eqref{eq:accept}, we reuse their indices. Otherwise, we retrieve a new set of representatives.

Let $N$ be the total number of tokens entering global attention, including $N_s$ special tokens. Let $M_Q$ be the number of queries retained for patch tokens and $M_K$ the number of key--value pairs after merging. For each head, dense global attention computes $N^2$ attention scores. MoDiR computes $M_QM_K$ scores for the retained queries of patch tokens and $N_sN$ for those of special tokens. Key and value projections still run at all positions. The end-to-end runtime includes geometry preparation, representative retrieval, and feature fitting. Appendix~\ref{app:d4rt} describes the use of feature fitting to compress decoder memory for 3D point tracking.

\section{Experiments}
\begin{table*}[!t]
\centering
\caption{Point-cloud reconstruction and inference efficiency on ScanNet30. CD is the sum of accuracy and completeness distances, and NC is normal consistency. Spd. includes online preparation. Mem. is peak allocated GPU memory (GiB) during the warmed-up main forward. OOM indicates that all 30 sequences at the given length exceeded GPU memory. Gray rows indicate the baseline models. Bold and underline mark the best and second-best distinct values among accelerated methods, with ties at the displayed precision.}
\label{tab:scannet}
\includegraphics[clip,width=\textwidth]{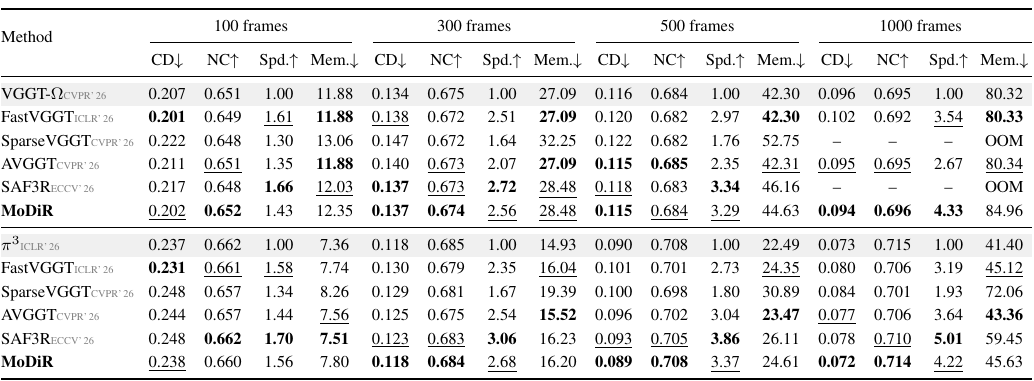}
\end{table*}

\begin{table*}[!t]
\centering
\caption{Depth and camera pose estimation on ScanNet30, using the same runs as Tab.~\ref{tab:scannet}. Rel denotes AbsRel, and A3 and A30 denote AUC@3 and AUC@30. $\delta_{1.25}$, A3, and A30 are percentages. Gray rows indicate the baseline models. Bold and underline mark the best and second-best distinct values among accelerated methods, with ties at the displayed precision.}
\label{tab:depth}
\includegraphics[clip,width=\textwidth]{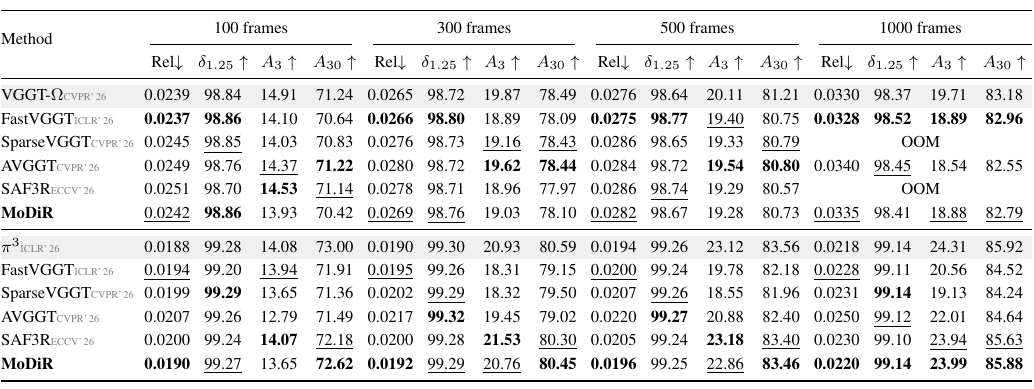}
\end{table*}

\begin{table*}[!t]
\centering
\caption{Geometry estimation with \om{} across datasets at 300 frames. TUM-Dynamics, NRGBD, and 7Scenes contribute 8, 9, and 18 sequences, respectively. Metrics follow Tabs.~\ref{tab:scannet} and \ref{tab:depth}. Gray rows indicate the baseline models. Bold and underline mark the best and second-best distinct values among accelerated methods, with ties at the displayed precision.}
\label{tab:cross}
\includegraphics[clip,width=.92\textwidth]{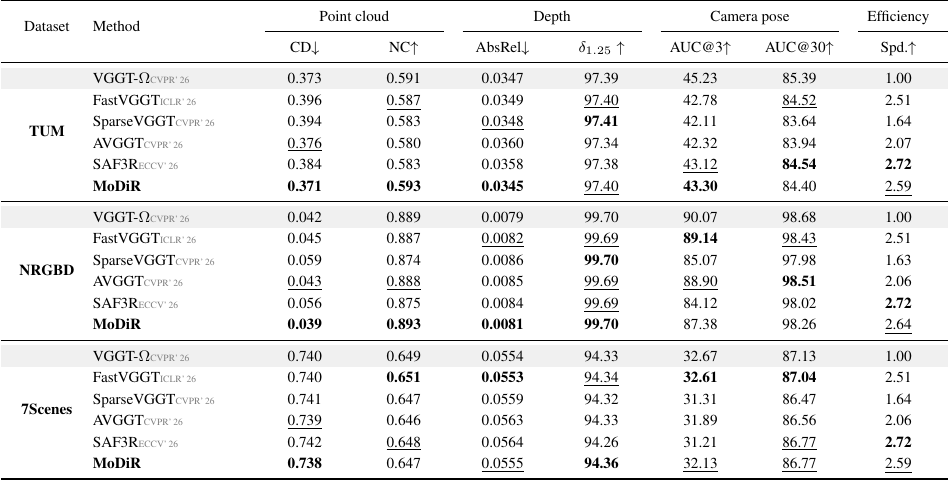}
\end{table*}

\begin{table}[!tb]
\caption{OpenD4RT on 50 clips per dataset. APD (\%) and EPE (dataset units) use the dynamic subset, available on 34 Waymo and 50 Kubric clips. Spd. uses all clips and is relative to OpenD4RT. Opt. denotes runtime optimizations with dense memory. FastVGGT (decoder) adapts FastVGGT to the OpenD4RT decoder. Gray rows mark the OpenD4RT baseline. Bold and underline identify the best and second-best compressed methods, respectively.}\label{tab:d4rt}
\centering\small\setlength{\tabcolsep}{2.35pt}
\begin{tabular}{llccc}\toprule
Data & Method & APD$\uparrow$ & EPE$\downarrow$ & Spd.$\uparrow$ \\
\midrule
\rowcolor{black!8} Waymo & OpenD4RT & 0.8858 & 10.7190 & 1.000 \\
 & OpenD4RT (opt.) & 0.8858 & 10.7190 & 1.52 \\
 & FastVGGT (decoder)\paperVenue{ICLR}{26} & \underline{0.9947} & \underline{10.8981} & \underline{1.783} \\
 & MoDiR & \textbf{1.0095} & \textbf{10.6617} & \textbf{1.788} \\
\midrule
\rowcolor{black!8} Kubric & OpenD4RT & 29.5414 & 2.0296 & 1.000 \\
 & OpenD4RT (opt.) & 29.5414 & 2.0296 & 1.54 \\
 & FastVGGT (decoder)\paperVenue{ICLR}{26} & \underline{29.3893} & \underline{2.0356} & \textbf{1.818} \\
 & MoDiR & \textbf{29.4210} & \textbf{2.0341} & \underline{1.814} \\
\bottomrule\end{tabular}
\end{table}

\begin{table}[!t]
\centering
\captionsetup{hypcap=false}
\caption{Component controls on eight 300-frame TUM-Dynamics sequences with \om{}. $K$ is the maximum number of representatives per fit. MoDiR uses spatial and temporal representatives with $K=4$. $\delta_{1.25}$ and AUC@30 are percentages. FPS includes online preparation. The gray row marks the baseline. Bold and underline mark the best and second-best distinct values among compressed configurations. Ties share highlighting. Full metrics are in Table~\ref{tab:component-full}.}
\label{tab:progressive}
\small\setlength{\tabcolsep}{3pt}
\begin{tabular}{@{}lcccc@{}}\toprule
Configuration & CD$\downarrow$ & $\delta_{1.25}\uparrow$ & AUC@30$\uparrow$ & FPS$\uparrow$ \\
\midrule
\rowcolor{black!8} \om{}\paperVenue{CVPR}{26} & 0.3733 & 97.39 & 85.39 & 11.03 \\
Spatial only, $K=1$ & 0.3767 & \underline{97.39} & 83.72 & 25.92 \\
Spatial + temporal, $K=1$ & 0.3779 & \underline{97.39} & 83.81 & 24.93 \\
Spatial only, $K=4$ & \underline{0.3752} & \textbf{97.40} & \underline{84.12} & \textbf{28.76} \\
MoDiR, $K=4$ & \textbf{0.3715} & \textbf{97.40} & \textbf{84.40} & \underline{28.59} \\
\bottomrule
\end{tabular}

\end{table}
\begin{table*}[!t]
\centering
\caption{Example results for dynamic-region depth estimation and throughput on Kubric and PointOdyssey~\cite{zheng2023pointodyssey}. The two selection rules use matched dictionary sizes. AbsRel and $\delta_{1.25}$ are reported as percentages. FPS includes online preparation.}
\label{tab:dynamic-examples}
\small
\setlength{\tabcolsep}{3.5pt}
\begin{tabular}{@{}lcl*{6}{>{\centering\arraybackslash}m{38pt}}@{}}
\toprule
 & & & \multicolumn{3}{c}{Selection by feature fitting error} & \multicolumn{3}{c}{Motion-aware selection} \\
\cmidrule(lr){4-6}\cmidrule(lr){7-9}
Dataset & Frames & Backbone
 & AbsRel$\downarrow$ & $\delta_{1.25}\uparrow$ & FPS$\uparrow$
 & AbsRel$\downarrow$ & $\delta_{1.25}\uparrow$ & FPS$\uparrow$ \\
\midrule
Kubric & 24 & \om{}\paperVenue{CVPR}{26} & 6.314 & 88.047 & 28.79 & 6.156 & 89.619 & 30.90 \\
 &  & \pii{}\paperVenue{ICLR}{26} & 3.100 & 99.659 & 39.62 & 3.010 & 99.665 & 40.02 \\
\midrule
PointOdyssey & 64 & \om{}\paperVenue{CVPR}{26} & 3.838 & 97.251 & 32.38 & 3.829 & 97.265 & 35.46 \\
 &  & \pii{}\paperVenue{ICLR}{26} & 12.550 & 97.146 & 54.61 & 12.494 & 97.160 & 57.59 \\
\bottomrule
\end{tabular}

\end{table*}

\subsection{Implementation Details}
\paragraph{Backbones and Baselines.}
We evaluate geometry estimation on \om{}~\cite{r20} and \pii{}~\cite{r22}. For 3D point tracking, we use feature fitting to compress OpenD4RT's decoder memory and retain all task queries. OpenD4RT is an unofficial implementation of D4RT~\cite{r25}. Appendix~\ref{app:d4rt} identifies the software revision and checkpoint. For each geometry backbone, we compare with dense inference and four acceleration methods, FastVGGT~\cite{r10}, SparseVGGT~\cite{r18}, AVGGT~\cite{r15}, and SAF3R~\cite{r5}. All comparisons use the same frozen checkpoint, inputs, and prediction heads within each backbone.

\paragraph{Datasets and Evaluation Metrics.}
We evaluate point-cloud reconstruction, depth estimation, and camera pose estimation on 30 ScanNet~\cite{r4} sequences at 100, 300, 500, and 1,000 frames, and at 300 frames on TUM-Dynamics~\cite{r13} (8 sequences), NRGBD~\cite{r1} (9), and 7Scenes~\cite{r11} (18). We report Chamfer distance (CD) and normal consistency (NC)~\cite{r12} after aligning each point cloud with a single similarity transform~\cite{r17} estimated from camera centers. We measure depth with absolute relative error (AbsRel) and $\delta_{1.25}$~\cite{r20} after scale and shift alignment per sequence. For camera pose, we report AUC at $3^\circ$ and $30^\circ$ over pairwise pose errors. Each sequence receives equal weight within its dataset.

\paragraph{Runtime and Memory.}
All experiments use one NVIDIA RTX PRO 6000 Blackwell GPU (96 GB). For geometry estimation, we time warmed-up online preparation, initial indexing and installation, and the mean of three synchronized forward passes. We charge preparation and setup once per sequence and report speedup relative to dense inference on the same backbone. GPU memory is the peak allocated memory during the warmed-up main forward. Appendix~\ref{app:protocol} details the timing and evaluation protocols.

\subsection{Geometry Estimation}
\paragraph{Point Cloud Reconstruction.}
On ScanNet, MoDiR achieves larger speedups as the input sequences grow longer on both backbones (Tab.~\ref{tab:scannet}). At 1,000 frames, MoDiR achieves $4.33\times$ and $4.22\times$ speedups over dense inference on \om{} and \pii{}, respectively, with slightly lower CD on both backbones (0.094 vs.\ 0.096 and 0.072 vs.\ 0.073, respectively). On \om{}, SparseVGGT and SAF3R run out of GPU memory on all 30 sequences at this length. MoDiR completes all 30 sequences, with a reported peak GPU memory allocation of 84.96~GiB during the main forward pass after warm-up. Overall, MoDiR substantially accelerates long-sequence reconstruction while maintaining comparable point-cloud quality on both backbones.

\paragraph{Depth and Camera Pose Estimation.}
On the same 1,000-frame sequences, MoDiR achieves an AbsRel of 0.0335 and an AUC@30 of 82.79\% on \om{}, and 0.0220 and 85.88\% on \pii{} (Tab.~\ref{tab:depth}). On \pii{}, these are the best results among the accelerated methods. Across the two backbones, AUC@30 decreases by at most 0.39 percentage points relative to dense inference. MoDiR thus retains competitive depth and pose accuracy at the speedups reported in Tab.~\ref{tab:scannet}.

\paragraph{Results Across Datasets.}
At 300 frames, MoDiR achieves $2.59$--$2.64\times$ speedups on \om{} and the lowest CD among the accelerated methods on TUM-Dynamics, NRGBD, and 7Scenes (Tab.~\ref{tab:cross}). On NRGBD, MoDiR reduces CD from 0.042 to 0.039 relative to dense inference, while AUC@3 decreases from 90.07\% to 87.38\%. Across the three datasets, the clearest gains are in point-cloud reconstruction, where MoDiR improves both reconstruction accuracy and inference efficiency on \om{}.

\subsection{3D Point Tracking}
\paragraph{Evaluation Protocol.}
We evaluate decoder-memory compression on 50 Waymo~\cite{r14} and 50 Kubric~\cite{r7} clips from the 4RC evaluation set~\cite{fourrc}, each with 24 consecutive frames, using OpenD4RT's 48CLIP checkpoint. All configurations share the same encoder and task queries. For visible points in the dynamic subset, we report endpoint error (EPE)~\cite{fourrc} and the percentage of points within a distance threshold, averaged over four fixed thresholds (APD)~\cite{tapvid3d}. Throughput is measured on all clips and includes encoding, preparation, decoding, and readback. Appendix~\ref{app:d4rt} specifies the subset and alignment protocol.

\paragraph{Quantitative Results.}
On Waymo and Kubric, MoDiR achieves $1.788\times$ and $1.814\times$ the throughput of unmodified OpenD4RT, respectively (Tab.~\ref{tab:d4rt}). Relative to the optimized baseline with dense decoder memory, the throughput gains are about 17\% and 18\%. MoDiR also achieves the highest APD and lowest EPE among the compressed variants on both datasets. These results show that decoder-memory compression improves throughput while retaining competitive tracking accuracy across both datasets.

\subsection{Ablation Studies}\label{sec:ablations}
\paragraph{End-to-End Ablation.}
With $K=4$, MoDiR achieves a CD of 0.3715 and an AUC@30 of 84.40\%, compared with 0.3752 and 84.12\% using spatial representatives alone at the same $K$ (Tab.~\ref{tab:progressive}). Throughput is similar at 28.59 and 28.76~FPS, respectively. At $K=1$, adding temporal representatives increases CD. The benefit of temporal representatives therefore depends on the maximum number of representatives per fit. At $K=4$, they improve reconstruction and pose accuracy with little change in throughput.

\paragraph{Representative Selection.}
With the number of added representatives fixed, MoDiR's motion-aware selection achieves the lowest relative attention output error on the four TUM-Dynamics sequences, at 6.985\% compared with 6.999\% for selection by feature fitting error (Tab.~\ref{tab:motion-matched}). The corresponding fit failure rates are 3.484\% and 3.412\%. Motion cues therefore yield a small improvement in output approximation at the same dictionary size, with a slightly higher fit failure rate.

\begin{figure}[!tb]
\centering
\includegraphics[width=.72\linewidth]{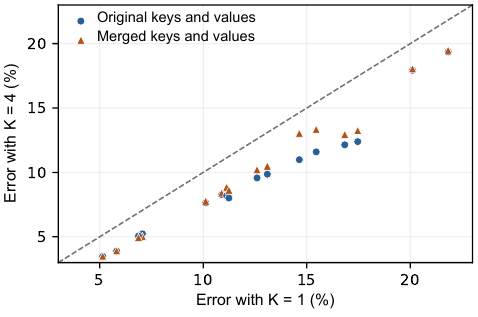}
\caption{Relative attention output error with one versus up to four representatives per fit. At $K=4$, we compare original and merged keys and values. Both comparisons use the same one-representative reference with original keys and values, drawn from the same candidate set. Each marker represents one backbone, sequence, and layer.}\label{fig:recovery}
\end{figure}

\paragraph{Attention Output Approximation.}
On two TUM-Dynamics sequences, we hold the layer inputs, queries, and candidate set fixed and evaluate tokens accepted for merging in the $K=4$ configuration (Fig.~\ref{fig:recovery}). With original keys and values, using up to four representatives instead of one reduces relative attention output error by 10.8--33.1\%. With merged keys and values, the reduction relative to the same single-representative reference is 10.2--32.5\%. The reduction in approximation error persists when keys and values are merged.

\paragraph{Depth Estimation in Dynamic Regions.}
We compare motion-aware selection with selection by feature fitting error at matched dictionary sizes, using the same fitting procedure (Tab.~\ref{tab:dynamic-examples}). On the Kubric example with \om{}, motion-aware selection reduces AbsRel from 6.314\% to 6.156\% and increases $\delta_{1.25}$ by 1.572 percentage points. Appendix~\ref{app:dynamic-examples} details the evaluation protocol. Across the reported examples from Kubric and PointOdyssey, motion-aware selection improves dynamic-region depth accuracy and throughput at the same dictionary size.\par

\section{Conclusion}
We present MoDiR, a motion-aware token reduction method that uses scene motion to guide representative selection in feed-forward 4D reconstruction. MoDiR builds a dictionary of spatial and temporal representatives and reuses token-specific fitting weights to approximate global attention outputs. Experiments show competitive accuracy in geometry estimation and 3D point tracking, with up to a $4.3\times$ inference speedup for long-sequence geometry estimation.

{\small\bibliographystyle{ieeenat_fullname}\bibliography{main}}
\clearpage
\appendix
\numberwithin{equation}{section}
\renewcommand{\theHequation}{appendix.\Alph{section}.\arabic{equation}}
\twocolumn[{\centering\Large\bfseries Supplementary Material\par\bigskip}]
\section{Algorithmic Procedure}\label{app:algorithm}
Algorithm~\ref{alg:modir} summarizes how MoDiR selects representatives and approximates attention outputs in \om{} and \pii{}. We prepare motion cues once per sequence, update temporal representatives periodically, and refit token-specific weights at each compressed global attention layer.

\begin{algorithm}[H]
\caption{MoDiR for feed-forward reconstruction}\label{alg:modir}
\small\raggedright
\algrenewcommand\algorithmicindent{0.9em}
\begin{algorithmic}[1]
\Require Time-ordered frames $I_{1:T}$, frozen backbone, maximum number of representatives per fit $K$
\Ensure Reconstruction predictions
\State Prepare motion cues for the input sequence
\State Sample spatial representatives $\mathcal S$ and initialize the temporal representative set $\mathcal E\gets\varnothing$
\For{each compressed global attention layer $l$}
  \State When an update is due, use motion cues to identify candidates and add those poorly represented by $\mathcal S\cup\mathcal E$ to $\mathcal E$ within the remaining budget
  \State Form $\mathcal D^l\gets\mathcal S\cup\mathcal E$. Fit each remaining patch token feature at this layer as a sparse convex combination of up to $K$ dictionary entries
  \State Compute attention outputs for retained queries. For each omitted query, reuse its fitting weights to combine the representatives' attention outputs
\EndFor
\end{algorithmic}
\end{algorithm}

\paragraph{OpenD4RT decoder.}
For 3D point tracking, we use only spatial representatives to compress the decoder memory and retain all task queries (Appendix~\ref{app:d4rt}).

\section{Implementation Details}\label{app:implementation}
\subsection{Native Observations and Initialization}
We select evenly spaced anchor frames at a rate of approximately one per 25 input frames, including the first and last frames. The frozen backbone jointly estimates 3D geometry on these anchors. To obtain one 3D position per patch, we backproject the predicted depth at the patch center for \om{} and average the predicted 3D points within the patch for \pii{}.

We propagate these 3D positions to the intervening frames using bidirectional optical flow between adjacent frames. Each patch receives one position estimate from the preceding anchor and another from the following anchor. We combine the estimates according to the temporal distance to each anchor and the flow confidence accumulated along each path. Confidence is higher when the backward flow more closely reverses the forward displacement. The resulting 3D positions initialize the test-time refinement.

\subsection{Test-Time Geometry Refinement}
The test-time training (TTT) block in Fig.~\ref{fig:method} refines the 3D positions propagated from the anchor frames. We freeze the backbone and update only a small adapter for each sequence. The adapter predicts a correction to each patch's initial position. Anchor predictions constrain the refined positions at anchor frames. Between adjacent frames, we encourage agreement between patches matched by flow, weighted by correspondence confidence. Before refinement, we use the initial geometry and optical flow to flag motion-sensitive and uncertain patches. Matches with a flagged target patch are excluded from this constraint.

After 12 optimization steps, we combine the refined position $p_i$ of patch $i$ with its normalized observation time $\bar t_i$ to form the geometry--time descriptor $g_i=(p_i,\eta\bar t_i)$, with $\eta=0.25$. We then recompute motion sensitivity from the refined positions and optical flow and use the resulting scores $s_i$ to select candidates for temporal representatives. 

\subsection{Motion Scoring and Temporal Representative Selection}
The spatial representatives $\mathcal S$ comprise all first-frame patches and one sampled patch per $2\times2$ region in each subsequent frame. To identify candidates for temporal representatives, we compute the motion sensitivity score
\begin{equation}
s_i=w_c(1-c_i)+w_m\hat m_i+w_h\hat h_i .
\end{equation}
Here $c_i$ measures the reliability of the flow correspondence, $m_i$ is the 3D position discrepancy between corresponding patches, and $h_i$ measures local variation in this discrepancy and in optical flow. Hats denote normalization to $[0,1]$, and $w_c,w_m,w_h$ are fixed weights. Thus an unreliable match can increase $s_i$ even when the measured 3D displacement is small.

Starting at the first compressed layer, we update temporal representatives every four compressed layers. We group motion-sensitive patches outside the dictionary by 3D position and four-frame intervals. For \om{}, the anchor coordinates used to initialize these positions are centered by their coordinatewise median, divided by the 75th-percentile radius, and clipped to $[-4,4]$. Each group $G$ proposes one candidate $r_G$ by Eq.~\eqref{eq:representative-score}. The three terms favor a patch whose feature $f_i$ is close to the group mean feature, whose descriptor $g_i$ is close to the group mean descriptor, and whose motion sensitivity $s_i$ is high.

At the first update, we fit each candidate's feature using the spatial representatives. At later updates, we also use temporal representatives selected in previous updates. A candidate that passes the fit test is already represented. We rank candidates that fail by the product of $s_i$ and the feature fitting error. We add the highest-ranked candidates to $\mathcal E$ within the remaining budget (the cumulative limit is 2\% of the input patch tokens, rounded to the nearest integer). The spatial and temporal representatives together form the dictionary $\mathcal D^l=\mathcal S\cup\mathcal E$ for the current layer.

\subsection{Representative Retrieval and Fitting}
For each patch token $i$ to be fitted, $S_i$ in Fig.~\ref{fig:method} indexes up to $K=4$ representatives from the current dictionary. We index the spatial representatives $\mathcal S$ once per sequence using features from the first compressed layer. We retrieve spatial candidates by feature similarity and proximity in geometry--time descriptor space, and temporal candidates by feature similarity. We remove duplicates and rank candidates by their current normalized feature similarity to token $i$. At a later compressed layer, we can reuse the same representative indices, but we refit the weights $a_{ij}$ because the token and representative features have changed. If a previous representative is unavailable or the new fit fails, we retrieve candidates again and repeat the fit.

When four representatives are retrieved, we evaluate all 15 nonempty subsets. For each subset, we fit nonnegative weights that sum to one (Eq.~\eqref{eq:fit}) and compare feasible fits by their squared feature error. We select the fit with the lowest error, using the first retrieved representative alone if it gives higher cosine similarity to the token feature. We then apply the fit test in Eq.~\eqref{eq:accept}. On both backbones, the minimum fitted-feature norm $\rho_{\min}$ is 0.8. The cosine threshold $\tau_l$ is 0.90, increasing to 0.94 at compressed layers 8, 10, 11, and 12. We number compressed layers from zero.

Patch tokens whose features fail the fit test retain their queries and receive directly computed outputs. At selected layers, \pii{} also retains queries for some tokens whose feature fits pass. At compressed layers 13, 14, 16, and 17, we retain additional queries with high fitting residuals to target retention rates of 95\%, 95\%, 45\%, and 55\%, respectively, measured over patch and special tokens together. If the queries already retained exceed the target, we keep them all. This adds query rows while preserving the dictionary size and the number of key/value pairs. \om{} uses no additional query protection.

\section{Feature Approximation with Multiple Representatives}\label{app:theory}
Let $f$ be a token feature and $D=[d_1,\ldots,d_m]$ a fixed set of candidate representatives. Denote by $E_K(f;D)$ the minimum squared error when approximating $f$ by a convex combination of at most $K$ representatives. Assigning weight one to any single representative remains feasible for every $K\geq1$, so
\begin{equation}
E_K(f;D)\leq E_1(f;D)=\min_j\|f-d_j\|_2^2.
\end{equation}
With two representatives, the approximation can lie anywhere on the line segment between their features. If that segment passes closer to $f$ than the nearest individual representative, the feature error is strictly smaller. Figure~\ref{fig:recovery} and Table~\ref{tab:merge-diagnostic} compare attention output approximations obtained with one or up to four representatives drawn from the same candidate set for each token. With the original keys and values, using up to four representatives yields lower relative attention output error than using one representative in all 16 comparisons.

\section{Analysis of Representative Selection and Attention Output Approximation}\label{app:diagnostics}
\paragraph{Observations on token merging.}
We applied the token merging procedure of FastVGGT~\cite{r10} and examined the 3D positions and attention outputs of the tokens it grouped together. Some merged pairs had highly similar image encoder features even though the corresponding patches were far apart in 3D. The pair in Fig.~\ref{fig:motivation}(a) has a feature cosine similarity of 0.975, yet its reference 3D positions are 0.224~m apart. We also observed substantial differences between the original attention outputs of tokens assigned to the same representative. Each row of Fig.~\ref{fig:motivation}(b) illustrates this observation at global attention layer 12 of \om{}. For two such tokens with projected outputs $o_i$ and $o_k$ before residual addition, the normalized output difference is
\begin{equation}d_o=\frac{\|o_i-o_k\|_2}{\sqrt{(\|o_i\|_2^2+\|o_k\|_2^2)/2}}.
\end{equation}

\subsection{Attention Output Approximation}
Figure~\ref{fig:recovery} and Table~\ref{tab:merge-diagnostic} report 16 comparisons across two backbones, two 300-frame TUM-Dynamics sequences, and four global attention layers per backbone. For each comparison, we fix the layer input and sample 4,096 tokens from those that pass the fit test at $K=4$. We evaluate all three configurations on these same tokens. The single-representative baseline uses the closest feature match from the same candidate set used for the four-representative fit.

The first two result columns use the original keys and values to compare one representative with up to four. For the third column, we keep the representatives and fitting weights from the $K=4$ configuration and recompute attention with merged keys and values. We measure the approximate outputs $\hat o_i$ against the original outputs $o_i$ after output projection and before residual addition. Only the original outputs propagate to the next layer. Over the sampled positions $I$, the relative attention output error is
\begin{equation}e=\sqrt{\frac{\sum_{i\in I}\|o_i-\hat o_i\|_2^2}{\max(\sum_{i\in I}\|o_i\|_2^2,10^{-20})}}.
\end{equation}
Even with merged keys and values, combining up to four representative outputs per token reduces relative attention output error by 10.2--32.5\% across the 16 comparisons relative to copying a single representative's output computed with the original keys and values.

\begin{table}[t]
\caption{Relative attention output error (\%) at fixed layer inputs, evaluated on tokens that pass the fit test at $K=4$. Rpy and Half denote \texttt{walking\_rpy} and \texttt{walking\_halfsphere}. $K$ is the maximum number of representatives per fit. The first two configurations use original keys and values (K/V). The third uses merged K/V.}\label{tab:merge-diagnostic}
\centering\small\setlength{\tabcolsep}{3.2pt}
\begin{tabular}{llcccc}\toprule
Backbone & Clip & Layer & $K=1$ & $K=4$ & \shortstack{$K=4$\strut\\Merged K/V\strut} \\
\midrule
\om{} & Rpy & 0 & 10.12 & 7.65 & 7.78 \\
 &  & 8 & 17.46 & 12.39 & 13.25 \\
 &  & 12 & 15.46 & 11.59 & 13.35 \\
 &  & 18 & 21.83 & 19.36 & 19.49 \\
\midrule
\om{} & Half & 0 & 10.89 & 8.28 & 8.41 \\
 &  & 8 & 16.83 & 12.14 & 12.95 \\
 &  & 12 & 14.64 & 10.99 & 13.04 \\
 &  & 18 & 20.11 & 17.93 & 18.06 \\
\midrule
\pii{} & Rpy & 0 & 5.82 & 3.90 & 3.93 \\
 &  & 8 & 12.60 & 9.57 & 10.21 \\
 &  & 12 & 11.13 & 8.19 & 8.85 \\
 &  & 17 & 7.06 & 5.23 & 5.05 \\
\midrule
\pii{} & Half & 0 & 5.14 & 3.49 & 3.51 \\
 &  & 8 & 13.10 & 9.86 & 10.49 \\
 &  & 12 & 11.24 & 8.01 & 8.63 \\
 &  & 17 & 6.88 & 5.06 & 4.95 \\
\bottomrule\end{tabular}\end{table}

\subsection{Effect of Representative Selection}
With the number of added representatives fixed, we compare random selection, selection by feature fitting error, and motion-aware selection, starting from the same spatial representatives. The fitting-error rule prioritizes patches whose features are poorly approximated by the spatial representatives. We also evaluate the spatial representatives alone. On four 300-frame TUM-Dynamics sequences with \pii{}, we evaluate layers 8, 12, and 17. At each layer of each sequence, all configurations use $K=4$ and the same 8,192 sampled tokens, none of which serves as a representative in any configuration.

We combine the representatives' original attention outputs using the fitted weights and score all sampled tokens, including those that fail the fit test. Motion-aware selection yields the lowest relative attention output error on all four sequences (Table~\ref{tab:motion-matched}). Its mean error is 6.985\%, compared with 6.999\% for selection by fitting error and 7.019\% for random selection.

\begin{table}[t]
\caption{Relative attention output error (\%) on four 300-frame TUM-Dynamics sequences with \pii{}. Random, fitting-error, and motion-aware selection add equal numbers of representatives. Spatial only is the reference without additions. Each sequence averages the relative $L_2$ error over three layers. The mean is taken over all 12 sequence--layer pairs.}\label{tab:motion-matched}
\centering\small\setlength{\tabcolsep}{3pt}
\begin{tabular}{@{}lcccc@{}}\toprule
Sequence & \shortstack{Spatial\\only} & Random & Fit error & Motion-aware \\
\midrule
walking\_xyz & 7.266 & 7.200 & 7.177 & \textbf{7.159} \\
walking\_halfsphere & 7.898 & 7.786 & 7.791 & \textbf{7.769} \\
sitting\_halfsphere & 7.185 & 7.156 & 7.124 & \textbf{7.115} \\
sitting\_xyz & 5.952 & 5.932 & 5.904 & \textbf{5.899} \\
\midrule
Mean & 7.075 & 7.019 & 6.999 & \textbf{6.985} \\
\bottomrule
\end{tabular}
\end{table}

\subsection{Depth Estimation in Dynamic Regions}\label{app:dynamic-examples}
We next examine how the choice of temporal representatives affects depth estimation in dynamic regions (Table~\ref{tab:dynamic-examples}). At each update, motion-aware selection and selection by feature fitting error add the same number of temporal representatives. Both use the same geometry preparation, representative retrieval, and output approximation procedure, with up to four representatives per fit.

Both Kubric and PointOdyssey provide ground-truth 3D point trajectories. We distinguish moving and static points by their displacement over time in a common reference frame and evaluate depth at the image locations of visible track points. For Kubric, we use the camera-space depth of each annotated point as the reference. For PointOdyssey, we read the reference depth from the supplied depth map at the projected location. We estimate one scale and shift per sequence using static points and apply the same alignment to evaluate depth at moving points. Across the four comparisons in Table~\ref{tab:dynamic-examples}, motion-aware selection yields lower AbsRel and higher $\delta_{1.25}$.

\section{Detailed Evaluation Protocol}\label{app:protocol}
\subsection{Geometry Evaluation Metrics}
We report camera pose AUC at $3^\circ$ and $30^\circ$, taking the larger of the relative rotation and translation-direction errors for each camera pair. For depth, we fit one scale and shift per sequence by minimizing absolute depth error over valid reference depths ($0<d<10$), then compute AbsRel and $\delta_{1.25}$. For point clouds, we estimate one similarity transform per sequence from the camera centers~\cite{r17}. We report Chamfer distance as the sum of the mean nearest-neighbor distances in both directions, and normal consistency as the mean absolute dot product of the corresponding normals across both directions.

\subsection{Evaluation Sequences and Baseline Settings}
ScanNet30 comprises the 30 ScanNet sequences with the most valid frames. We evaluate all eight TUM-Dynamics sequences and the complete test splits of 7Scenes (18 sequences) and NRGBD (nine sequences). We sample every other frame when possible and otherwise sample uniformly over the sequence.

FastVGGT uses a merge ratio of 0.9, with a protection ratio of 0.1 on \om{}. SparseVGGT uses average-pooled prediction with sparsity 0.5 and SpargeAttn kernels. We reimplemented AVGGT following the method described by Sun et al.~\cite{r15}. On both backbones, AVGGT uses a patch-token key/value subsampling factor of $\sigma=2$ and a fixed schedule for replacing early global attention layers with frame attention. We use the official SAF3R kernels. For offline calibration, we profile each backbone on 128 uniformly sampled frames from each of three 7Scenes training sequences. We include online index rebuilding in the reported timings and exclude offline calibration.

\subsection{Runtime and Memory}
We measure geometry inference on one NVIDIA RTX PRO 6000 Blackwell GPU (96 GB). End-to-end inference time includes online preparation and the main forward pass. For the spatial-only configuration, online preparation consists of spatial dictionary construction. Configurations with temporal representatives also include the time spent extracting motion cues.

For each sequence, we warm up preparation once and measure a second run on the same input. We also warm up the main forward pass once and average three synchronized runs. Throughput is the total number of frames divided by the total inference time across sequences. We measure peak allocated GPU memory during the main forward pass after warmup. The backbone revisions are \texttt{48b23c8ce72c} for \om{} and \texttt{9fa3ddb3f8d5} for \pii{}.

\subsection{3D Point Tracking with OpenD4RT}\label{app:d4rt}
We compress the decoder memory of OpenD4RT, an unofficial implementation of D4RT~\cite{r25}, while retaining all task queries. We fit each remaining memory token using up to four spatial representatives. Tokens that fail the fit test retain their own memory entries. All queries attend to the full memory in the first decoder layer and to the compressed memory in the remaining seven layers.

At the retained memory positions, we use the original projected keys and values. Each token that passes the fit test distributes unit weight across its representatives according to its fitted weights. We add the logarithm of the total weight represented by each memory entry to its attention logit. The FastVGGT decoder variant uses a single representative per token and averages keys and values according to the weights assigned to them.

We use revision \texttt{403290a6e7ea} of the \href{https://github.com/Lijiaxin0111/Open-d4rt}{\texttt{Lijiaxin0111/Open-d4rt}} repository with the \texttt{48CLIP}, \texttt{9Mix}, \texttt{NoCropAUG} checkpoint variant. Table~\ref{tab:d4rt} reports results on 50 Waymo clips and 50 Kubric clips of 24 consecutive frames each. All configurations use the same input frames and task queries.

For Waymo and Kubric, we compare predicted and ground-truth 3D tracks in their common reference frame~\cite{fourrc}. We identify dynamic tracks by summing the distances between consecutive ground-truth positions and applying a threshold of 0.01 in dataset units. This yields 34 Waymo clips and 50 Kubric clips with dynamic tracks. We evaluate EPE and APD on visible points of the dynamic tracks after sequence-level median-scale alignment on those tracks. EPE is the mean 3D endpoint error. APD averages the percentage of points within each of four distance thresholds (0.1, 0.3, 0.5, and 1.0 dataset units). For sparse tracks stored in camera coordinates, we use the supplied extrinsics to transform them into the first camera's coordinate frame. Where extrinsics are unavailable, we evaluate predictions in the camera coordinates of each frame.

We measure throughput on all 50 clips per dataset in Table~\ref{tab:d4rt}. Each timed forward pass includes video encoding, decoder preparation, decoding all task queries, and transfer of the outputs to the CPU. We average three synchronized runs after one warmup. Speedups in Tables~\ref{tab:d4rt} and~\ref{tab:archive} are relative to unmodified OpenD4RT. The optimized dense-memory baseline in Table~\ref{tab:d4rt} measures the gain from runtime optimizations before memory compression.

\section{Additional Results}\label{app:additional}

\paragraph{Geometry estimation across datasets.}
Table~\ref{tab:pi3} extends the 300-frame evaluation in Table~\ref{tab:cross} to \pii{}. Using the same configuration on TUM-Dynamics, NRGBD, and 7Scenes, MoDiR achieves a $2.67$--$2.74\times$ speedup over dense inference.

\paragraph{3D point tracking across datasets.}
Across the six evaluation subsets in Table~\ref{tab:archive}, MoDiR achieves $1.20$--$1.81\times$ the throughput of unmodified OpenD4RT. On each subset, dynamic-track APD differs from the OpenD4RT baseline by less than 0.35 percentage points.

\begin{table*}[p]
\centering
\begin{minipage}{\textwidth}
\centering
\captionof{table}{Geometry estimation with \pii{} across datasets at 300 frames. Point-cloud distances use the coordinate units of each dataset. Rel denotes AbsRel, and A3 and A30 denote AUC@3 and AUC@30. Memory is in GiB. Gray rows indicate the baseline models. Bold and underline mark the best and second-best distinct values among accelerated methods, with ties at the displayed precision.}
\label{tab:pi3}
\includegraphics[clip,width=\linewidth]{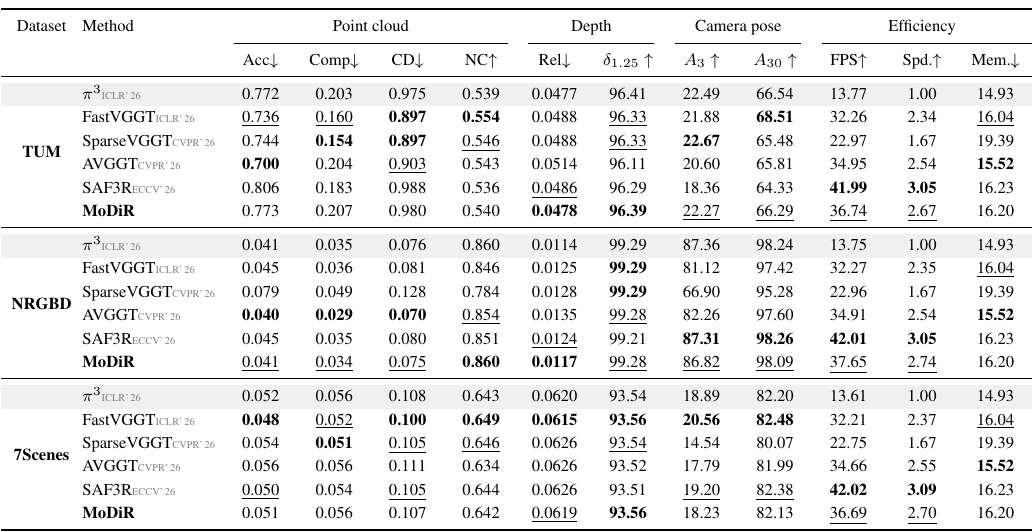}
\end{minipage}

\par\vspace{2\baselineskip}
\begin{minipage}{\textwidth}
\captionof{table}{3D point tracking with OpenD4RT across six evaluation subsets. APD (\%) and EPE (dataset units) are evaluated on dynamic tracks. Throughput includes all completed clips. Speedups are relative to OpenD4RT (gray). FastVGGT (decoder) adapts token merging to the OpenD4RT decoder. Bold and underline mark the best and second-best distinct values among the compressed methods.}\label{tab:archive}
\centering\small
\setlength{\tabcolsep}{13.65pt}
\begin{tabular}{llccccc}\toprule
Dataset & Method & APD$\uparrow$ & EPE$\downarrow$ & FPS$\uparrow$ & Spd.$\uparrow$ & Mem. (GiB)$\downarrow$\\
\midrule
\rowcolor{black!8} Waymo & OpenD4RT & 0.8858 & 10.7190 & 11.50 & 1.000 & 10.59 \\
 & FastVGGT (decoder)\paperVenue{ICLR}{26} & \underline{0.9947} & \underline{10.8981} & \underline{20.51} & \underline{1.783} & \textbf{4.67} \\
 & MoDiR & \textbf{1.0095} & \textbf{10.6617} & \textbf{20.56} & \textbf{1.788} & \underline{4.80} \\
\midrule
\rowcolor{black!8} Kubric & OpenD4RT & 29.5414 & 2.0296 & 1.55 & 1.000 & 10.80 \\
 & FastVGGT (decoder)\paperVenue{ICLR}{26} & \underline{29.3893} & \underline{2.0356} & \textbf{2.81} & \textbf{1.818} & \textbf{5.00} \\
 & MoDiR & \textbf{29.4210} & \textbf{2.0341} & \textbf{2.81} & \underline{1.814} & \underline{5.16} \\
\midrule
\rowcolor{black!8} po\_mini & OpenD4RT & 75.8817 & 0.2463 & 126.09 & 1.000 & 10.59 \\
 & FastVGGT (decoder)\paperVenue{ICLR}{26} & \underline{75.7829} & \underline{0.2459} & \underline{204.15} & \underline{1.619} & \textbf{4.78} \\
 & MoDiR & \textbf{75.9010} & \textbf{0.2427} & \textbf{207.44} & \textbf{1.645} & \underline{4.89} \\
\midrule
\rowcolor{black!8} ds\_mini & OpenD4RT & 75.2372 & 0.2753 & 127.84 & 1.000 & 10.59 \\
 & FastVGGT (decoder)\paperVenue{ICLR}{26} & \underline{75.1472} & \underline{0.2762} & \underline{206.36} & \underline{1.614} & \underline{4.78} \\
 & MoDiR & \textbf{75.1710} & \textbf{0.2736} & \textbf{209.70} & \textbf{1.640} & \textbf{4.77} \\
\midrule
\rowcolor{black!8} adt\_mini & OpenD4RT & 76.2041 & 0.2881 & 228.93 & 1.000 & 10.58 \\
 & FastVGGT (decoder)\paperVenue{ICLR}{26} & \textbf{76.1760} & \textbf{0.2884} & \underline{319.24} & \underline{1.394} & \textbf{4.78} \\
 & MoDiR & \underline{76.0211} & \underline{0.2930} & \textbf{325.72} & \textbf{1.423} & \underline{4.85} \\
\midrule
\rowcolor{black!8} pstudio\_mini & OpenD4RT & 81.2916 & 0.1632 & 363.81 & 1.000 & 10.58 \\
 & FastVGGT (decoder)\paperVenue{ICLR}{26} & \textbf{80.9688} & \underline{0.1662} & \underline{433.42} & \underline{1.191} & \textbf{4.76} \\
 & MoDiR & \underline{80.9498} & \textbf{0.1659} & \textbf{437.22} & \textbf{1.202} & \underline{4.79} \\
\bottomrule
\end{tabular}
\end{minipage}

\end{table*}
\clearpage

\twocolumn[{
\begin{minipage}{\textwidth}
\centering
\captionof{table}{Full metrics for the controls in Table~\ref{tab:progressive} on the same eight 300-frame TUM-Dynamics sequences with \om{}. MoDiR uses spatial and temporal representatives with $K=4$. AUC and $\delta_{1.25}$ are percentages. FPS includes spatial dictionary construction for all component configurations and motion cue preparation for those using temporal representatives. Gray marks the baseline. Bold and underline mark the best and second-best distinct values among compressed configurations. Ties share highlighting.}
\label{tab:component-full}
\small
\begin{tabular}{lccccccc}\toprule
Configuration & AUC@3$\uparrow$ & AUC@30$\uparrow$ & AbsRel$\downarrow$ & $\delta_{1.25}\uparrow$ & CD$\downarrow$ & NC$\uparrow$ & FPS$\uparrow$ \\
\midrule
\rowcolor{black!8} \om{}\paperVenue{CVPR}{26} & 45.23 & 85.39 & 0.0347 & 97.39 & 0.3733 & 0.5905 & 11.03 \\
Spatial only, $K=1$ & 43.38 & 83.72 & \underline{0.0348} & \underline{97.39} & 0.3767 & 0.5918 & 25.92 \\
Spatial + temporal, $K=1$ & \textbf{43.52} & 83.81 & \underline{0.0348} & \underline{97.39} & 0.3779 & 0.5924 & 24.93 \\
Spatial only, $K=4$ & \underline{43.44} & \underline{84.12} & \textbf{0.0345} & \textbf{97.40} & \underline{0.3752} & \underline{0.5925} & \textbf{28.76} \\
MoDiR, $K=4$ & 43.30 & \textbf{84.40} & \textbf{0.0345} & \textbf{97.40} & \textbf{0.3715} & \textbf{0.5934} & \underline{28.59} \\
\bottomrule
\end{tabular}

\end{minipage}

\par\medskip
\begin{minipage}{\textwidth}
\begin{minipage}{\textwidth}
\centering
\includegraphics[width=.69\linewidth]{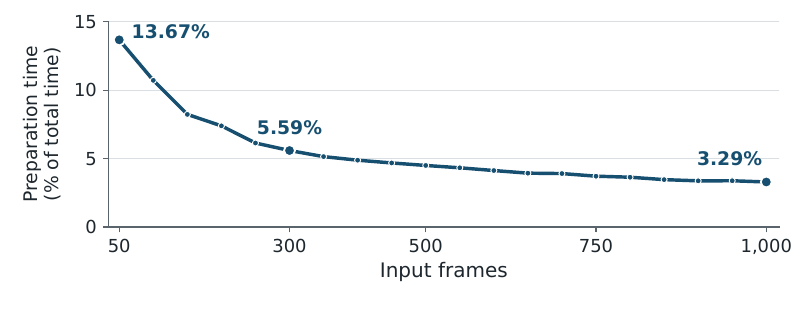}
\captionof{figure}{Sequence-level motion preparation time as a percentage of end-to-end inference time for \pii{} on ScanNet30 (RTX PRO 6000).}
\label{fig:motion-prep-share}
\end{minipage}
\end{minipage}

\par\smallskip
}]

\paragraph{Effect of temporal representatives.}
Table~\ref{tab:component-full} reports the full results of the component study in Table~\ref{tab:progressive} on eight 300-frame TUM-Dynamics sequences with \om{}. At $K=4$, adding temporal representatives reduces CD from 0.3752 to 0.3715 and raises AUC@30 from 84.12\% to 84.40\%. AUC@3 decreases by 0.14 percentage points, and throughput changes from 28.76 to 28.59 FPS.

\newpage
\paragraph{Motion preparation time.}
We measure motion preparation time for MoDiR with \pii{} at 20 input lengths on ScanNet30 (Fig.~\ref{fig:motion-prep-share}). Motion preparation accounts for 13.67\% of end-to-end inference time at 50 frames, falling to 5.59\% at 300 frames and 3.29\% at 1,000 frames. In a separate comparison at matched dictionary sizes, motion-aware selection improves both dynamic-region depth metrics in all four comparisons in Table~\ref{tab:dynamic-examples}. The largest gain in $\delta_{1.25}$ occurs on Kubric with \om{}, an increase of 1.57 percentage points over selection by feature fitting error. These results show that using motion cues to select temporal representatives helps preserve reconstruction accuracy with little additional inference time.

\clearpage
\begin{figure*}[p]
\centering
\includegraphics[width=.96\textwidth,height=.82\textheight,keepaspectratio]{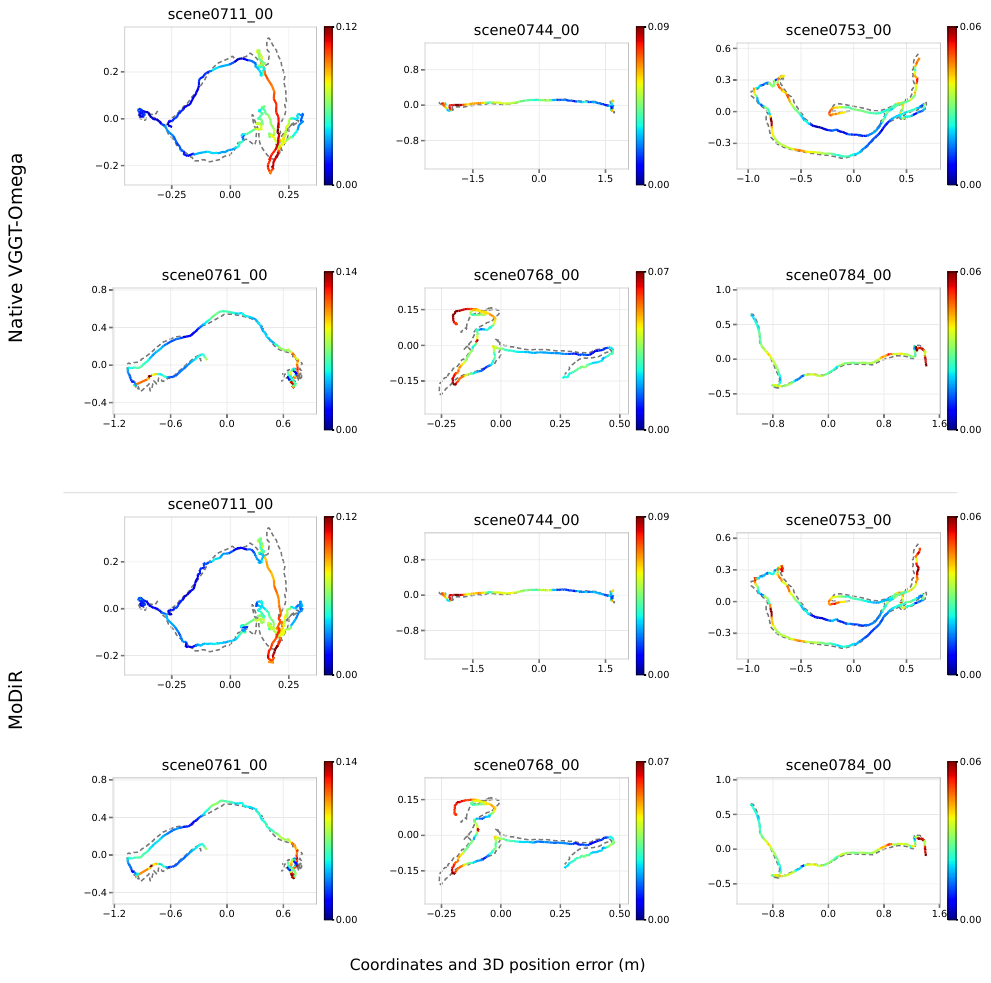}
\caption{\textbf{ScanNet30 camera trajectories.} Gray dashed lines show ground truth. Color indicates the 3D position error after sequence-level Sim(3) alignment. Each sequence uses one projection plane, axis extent, and color scale for both methods.}
\label{fig:scannet-trajectories}
\end{figure*}
\clearpage

\begin{figure*}[p]
\centering
\includegraphics[width=.96\textwidth,height=.82\textheight,keepaspectratio]{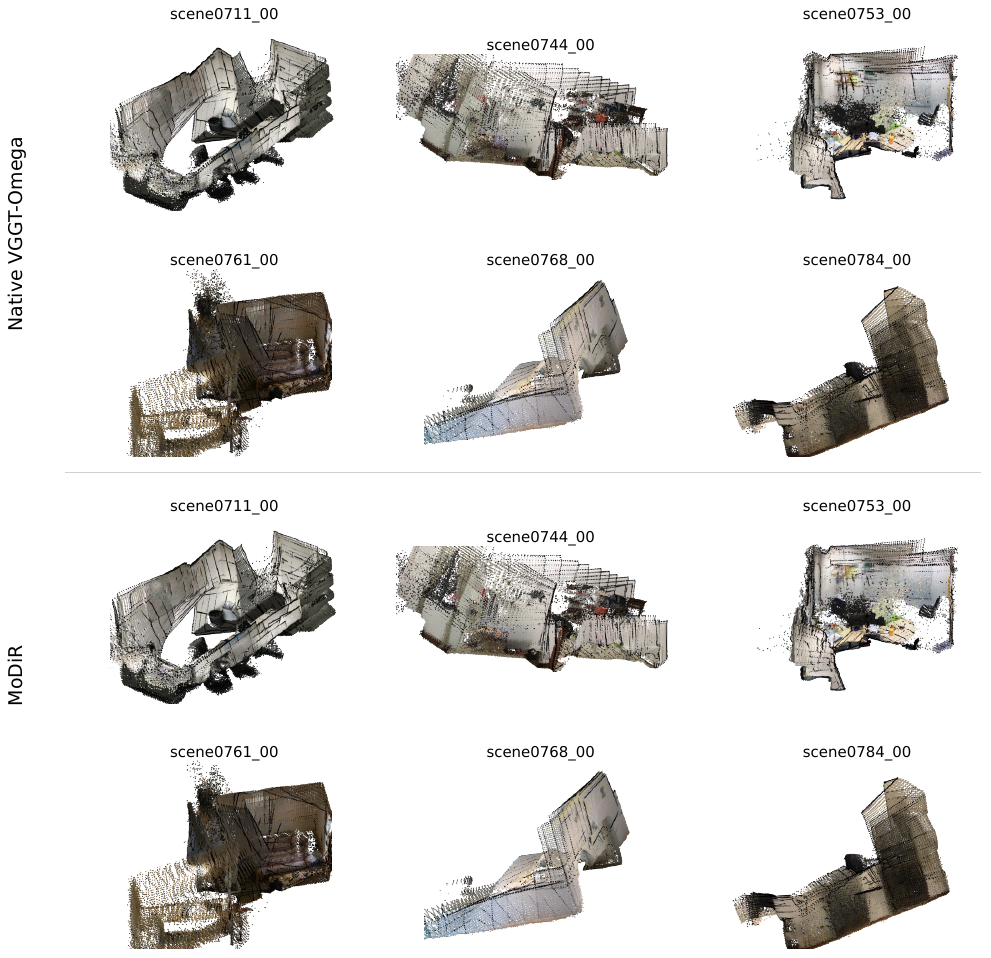}
\caption{\textbf{ScanNet30 RGB point clouds.} We backproject predicted depths using estimated cameras and color the points with input RGB. For display, we use the same 56 uniformly spaced frames and pixel grid for both methods. Each scene uses a single view and crop derived from its ground truth.}
\label{fig:scannet-clouds}
\end{figure*}

\end{document}